\documentclass[10pt,twocolumn,letterpaper]{article}

\usepackage{cvpr}
\usepackage{times}
\usepackage{epsfig}
\usepackage{graphicx}
\usepackage{amsmath}
\usepackage{amssymb}
\usepackage{epstopdf}
\usepackage{multicol}

\usepackage{xcolor}

\newcommand{\src}[1]{\textcolor{blue}{(#1)}}

\usepackage[breaklinks=true,bookmarks=false]{hyperref}

\cvprfinalcopy 

\def\cvprPaperID{15} 

\begin{document}

\title{SpliceRadar: A Learned Method For Blind Image Forensics}

\author{Aurobrata Ghosh$^1$ \qquad Zheng Zhong$^1$ \qquad Terrance E Boult$^2$ \qquad Maneesh Singh$^1$\\
$^1$Verisk AI, Verisk Analytics \qquad
$^2$Vision and Security Technology (VAST) Lab\\
{\tt\small \{aurobrata.ghosh, zheng.zhong, maneesh.singh\}@verisk.com} \quad
{\tt\small tboult@vast.uccs.edu}
}


\maketitle
\begin{figure*}[!h]
\begin{center}
   \includegraphics[width=0.95\linewidth]{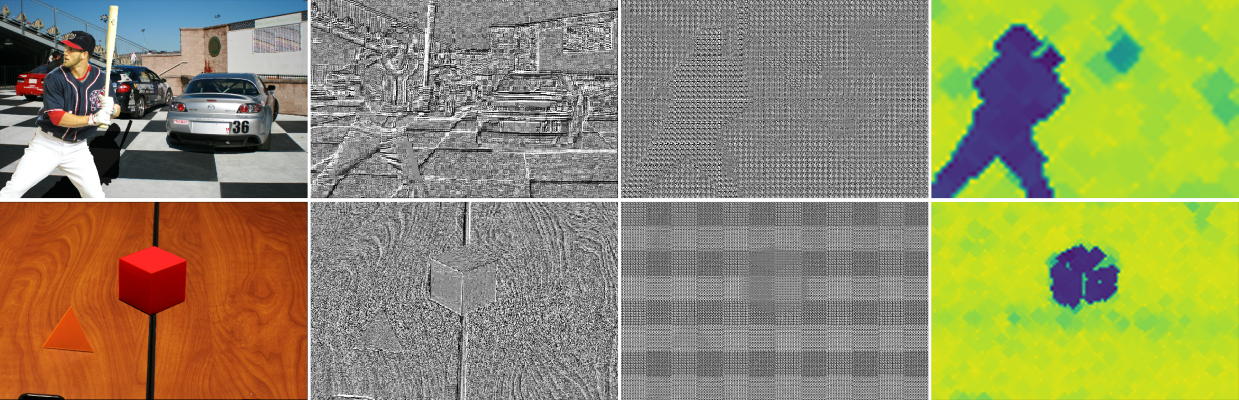}
\end{center}
   \caption{\textit{SpliceRadar} is able to learn low level features while suppressing 
   semantic-information which are image specific. This allows it to generalize well to new tampered datasets. 
   Two examples: 
   col-1: input image, col-2: sample of a learned rich filter (contains semantic-edges), col-3: final features    (semantic-edges suppressed), col-4: output heat map indicating tampered region.}
\label{fig:poster}
\end{figure*}

\begin{abstract}
 Detection and localization of image manipulations like splices are gaining in importance
 with the easy accessibility to image editing softwares.
 While detection generates a verdict for an image it provides no insight into the manipulation. 
 Localization helps explain a positive detection by identifying 
 the pixels of the image which have been tampered.
 We propose a deep learning based method for splice localization without prior 
 knowledge of a test image's camera-model. 
 It comprises a novel approach for learning rich filters and for suppressing image-edges. Additionally,
 we train our model on a surrogate task of camera model identification, which allows us to
 leverage large and widely available, unmanipulated, camera-tagged image databases. 
 During inference, we assume that the spliced and host regions come from different camera-models
 and we segment these regions using a Gaussian-mixture model.
 Experiments on three test databases demonstrate results on par 
 with and above the state-of-the-art and a good generalization ability to unknown datasets.

\end{abstract}

\section{Introduction}

``A picture is worth a thousand words''. A statement, which appeared in print in the early 1900s,
has become a ubiquitous part of our daily lives with the advance of camera technology. 
Ironically, however,
with media becoming digitized, this implicit trust is under attack. With the accessibility
of image editing softwares and wide diffusion of digital images over the internet,
anyone can easily create and distribute convincing fake pictures. These fakes have a significant impact on our lives: from the private, the social, to the legal.
It is imperative, therefore, to develop digital forensic tools capable of detecting such fakes. 


Typically, a fake well done hides its manipulations cleverly with the semantic contents 
of the image, therefore,
forensic algorithms inspect low-level statistics of images 
or inconsistencies therein to identify
manipulations. These include distinctive features stemming from the hardware and 
software of a particular camera make (or a post-processing step thereafter).
For example, at the 
lowest hardware level, the photo-response non-uniformity (PRNU) noise pattern is a digital noise
``fingerprint'' of a particular device and can be used for camera identification \cite{chen:08}. 
The colour filter array (CFA) and its interpolation algorithms are also particular 
to a device and can help discern
between cameras \cite{popescu:05}. At a higher level, the image compression format, \eg 
the popular JPEG format, can help determine single versus multiple compressions \cite{barni:17} 
or different device makes \cite{choi:06,qiao:17}. This is useful in the detection of digital edits and
localization of splices \cite{agarwal:17}.

Traditional image forensic algorithms have modelled discrepancies in one or multiple 
such statistics to
detect or localize splicing manipulations. 
Prior knowledge characterizing these discrepancies have been 
leveraged to design handcrafted features. 
The survey in \cite{zampoglou:16} compares the performances of a number of such algorithms.

Learned forensic approaches have recently gained popularity with the growing success of 
machine learning and deep learning. 
In \cite{cozzolino:17}, Cozzolino \etal recast hand designed
high pass filters, useful for extracting residual signatures, 
as a constrained CNN to learn the filters and residuals from a training dataset.
Zhou \etal \cite{zhou:18}, proposed a 
dual branch CNN, one learning from the image-semantics and the other learning from the image-noise,
to localize spliced regions. 
Huh \etal \cite{huh:18} (henceforth referred to as EXIF-SC), leveraged the EXIF metadata
to train a Siamese neural network to verify metadata consistency among patches of a test
image to localize manipulated pixels. In \cite{roessler:19}, R\"ossler \etal took on a new genre 
of forensic attacks -- state-of-the-art face manipulations including some 
created by deep neural networks -- and showed that learned CNNs outperformed traditional methods.
However, their success notwithstanding, deep learning approaches have typically 
shown vulnerability to 
generalizing to new datasets \cite{cozzolino:18,bappy:17,salloum:18}.

In this paper, we propose a novel, blind forensic
approach based on CNNs to localize spliced regions 
in an image without any prior knowledge of the source cameras.
We employ a new way to learn high pass ``rich'' filters 
and a novel probabilistic regularization based on 
mutual information to suppress semantic contents in the training images and 
learn low-level features of camera models. 
Our network
is trained for a surrogate task of source-camera identification, which allows us to use
large, widely available camera-tagged untampered images for training. 
Forgery localization is done by computing the low-level features of the image, which identifies
the signatures of multiple source camera models, and segmenting these regions using a
Gaussian mixture model. Preliminary results 
from a number of test databases: DSO-1 \cite{carvalho:13}, Nimble Challenge 2016
(NC16) and Nimble Challenge 2017 (NC17-dev1) \cite{medifor:17} show an improvement 
over the state-of-the-art.
Furthermore, since our training data is unrelated to the test datasets, it also demonstrates
good generalization ability.

In summary, the contributions in this paper are:
\begin{itemize}
    \item a new way to learn high pass rich filters using constrained CNNs
    that compute residuals, highlighting low-level information over the semantics
    of the image;
    \item  a novel probabilistic regularization based on mutual information, 
    which helps to suppress image-edges in the training data;
    \item experimental analysis showing up to $\sim4\%$ (points) improvement over the state-of-the-art on three 
    standard test datasets: DSO-1, NC16 and NC17-dev1.
\end{itemize}

\section{Related Work}

\noindent \textbf{Rich Filters:}
Spatial rich models for steganalysis \cite{fridrich:12}, proposed a large set of 
hand-engineered high pass filters, \textit{rich filters} (RFs), to extract local noise-like 
features from an image. By computing dependencies among neighbouring pixels,
these filters draw out residual information that highlights 
low-level statistics over the image-semantics. 
Rich filters have proven extremely effective in image forensics and have been
widely adopted by various state-of-the-art splice detection algorithms.
SpliceBuster (SB) \cite{cozzolino:15}, a blind splice detection algorithm, used one such 
fixed filter to separate camera features from the spliced and host regions. 
In \cite{zhou:18}, three fixed rich filters were used in the noise-branch to compute residuals 
along with a CNN to learn co-occurrence probabilities of the residuals as features to train 
a region proposal network to detect spliced regions.
Bayar and Stamm \cite{bayar:17,bayar:18}, proposed a constrained convolution
layer to learn RF-like features and a CNN to learn the co-occurrence probabilities from the data.
At every iteration they projected the weights of the constrained layer to satisfy 
$\mathbf{w}_k(0,0)=-1$ and $\sum_{m,n\neq0,0}\mathbf{w}_k(m,n)=1$,
where $\mathbf{w}_k(i,j)$ is the weight of the $k^{\textrm{th}}$ filter at position $(i,j)$.
The end-to-end trained network in \cite{bayar:18} was used to identify broad image-level 
manipulations like blurring and compression.
We also use learned RFs but propose a new constrained convolution layer and a 
different approach to applying the constraints.

\noindent
\textbf{Camera Identification:} Camera identification plays an important part in image forensics. 
Lukas \etal proposed a PRNU based camera identification algorithm in \cite{lukas:06} where they 
estimated nine reference noise patterns using wavelet denoising and averaging,
then matched the reference patterns to new images by correlation to determine the source camera.
CNNs were trained to compute features along with SVMs for source camera identification 
in \cite{bondi:17}.
Learned RFs from constrained convolution layers were used for camera identification in
\cite{bayar:17}.
Recently, Mayer and Stamm \cite{mayer:18}, trained a similar learned RF based CNN 
for a camera identification task, then used the output of the CNN as features to train a second
network for splice detection.
In \cite{bondi:17b}, Bondi \etal proposed a strategy similar to ours in spirit: a CNN as a feature
extractor to identify camera-models, patch based feature computation of a test image and 
clustering of the patch-features to localize spliced regions. 
However, it is fundamentally different from our proposed method.
Bondi \etal used regular convolutions and max-pooling in their CNN, which  
are typically used to learn high-level semantic structures of an image, therefore
biasing the CNN to learn semantic contents of the training data.
In this work, we propose to suppress the semantic contents of an
image to learn the distinguishing low-level features of a camera-model.
Additionally, the experiments in \cite{bondi:17b} are conducted on synthetic datasets with 
straightforward manipulations. 
In comparison, we demonstrate our method on multiple established test datasets with 
(series of) complex manipulations.



\section{Proposed Method}

We propose \textit{SpliceRadar} (SR), a deep learning approach for blind forgery localization. 
Our network has no prior knowledge of the source cameras of either the host or the spliced image
regions. Instead, it is trained to compute low-level features which can segregate camera-models.
A tampered region is localized by computing the features over the entire image and then segmenting
the feature-image using a Gaussian mixture model.

We train our network to differentiate camera-models instead of individual device instances. 
The learned features contain signatures of the entire image formation pipeline 
of a camera-model: from the hardware, the internal processing algorithms, to the compression.

Although challenging, we choose a blind localization strategy to improve the generalization
ability of our network. This is achieved by training the SpliceRadar 
network on a surrogate task of 
source camera-model identification, which allows us to leverage large and widely available 
camera-tagged image databases. It also allows us to avoid known manipulated datasets and risk 
the chance of over-specializing towards these.
Additionally, we train with a large number of camera models. This helps not only to generalize
better but also to boost our network's ability to segregate camera models. This ability to 
differentiate (even unknown) camera-models is of greater interest to us than the ability to
identify the models available during training.

\noindent
\textbf{Low-level features:} A key contribution in our design of SpliceRadar is its
ability to learn low-level features independent of the image-semantics. 
This is achieved in our architecture by a two-step process: 
residual information extraction and semantic-edge suppression.
The first layer of the network consists of a set of learned RFs
comparable to \cite{bayar:17,bayar:18}. These largely suppress the semantic contents of an input
patch from a colour-image by learning to compute residuals. 
However, since RFs are high-pass filters they 
also accentuate the semantic-edges present in the image (see Fig.~\ref{fig:poster}). 
Searching for patterns based on these will likely lead to learning
misleading image specific information that is not truly independent of the semantics.
This will result in our network learning information specific to the semantic 
contents of the training data, which would affect its generalization ability. 
Therefore,
after learning the spatial distribution of these residuals, we further suppress the remaining
semantic-edges by applying a probabilistic regularization. From these we learn 
a hundred-dimensional feature vector characteristic of a camera-model and
independent of the image-semantics. These features are used to 
drive a cross-entropy loss during training and for segmentation during forgery localization.


\noindent
\textbf{Learned RFs:} 
The first layer of our network computes residuals from learned filters that
resemble RFs in \cite{fridrich:12}.
We propose a novel way to do this using constrained convolutions that is
different from \cite{bayar:17,bayar:18}.
Developing along the lines of the original hand engineered RFs 
\cite{fridrich:12}, we define a residual to be the difference between a predicted value 
for a central pixel defined over its neighbourhood and the scaled value of the pixel. 
Therefore, from Eq.~1 in \cite{fridrich:12}, we propose the constrained convolution to learn
residuals as:
\begin{equation}
\mathcal{R}_{RF}^{(k)} = \mathbf{w}_k(0,0) + \sum_{m,n\neq0,0}\mathbf{w}_k(m,n) = 0,
\label{eq:RF}
\end{equation}
for the $k^{\textrm{th}}$ filter, 
where the support of the residuals is a $N\times N$ neighbourhood ($N=5$). 
The summation ensures that the 
predicted value and the pixel's value have opposite signs \cite{fridrich:12}.
Following the spirit of the original work, 
we propose to use a large bank of learned RFs, $k=1..64$, instead of only 3 learned RFs
like in \cite{bayar:17,bayar:18}.
These constraints are applied by including 
$\mathcal{R}_{RF}=(\sum_{k}(\mathcal{R}_{RF}^{(k)})^2)^{\frac{1}{2}}$ as a penalty
in the cost function.
This allows our network to learn suitable residuals for camera-model classification.

\begin{figure}[t!]
\begin{center}
   \includegraphics[width=1.\linewidth]{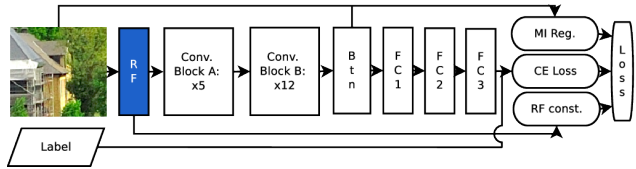}
\end{center}
   \caption{System architecture of SpliceRadar.
   }
\label{fig:network}
\end{figure}

\noindent
\textbf{System architecture:}
We propose an eighteen layer deep CNN that takes as input a 
$72\times72\times3$ RGB patch, and the camera-model label during training, 
as shown in Fig.~\ref{fig:network}. The first layer is
a constrained convolution layer with kernel size $5\times5\times3\times64$, 
producing 64 filters as described above. 
Convolution block A comprises of a convolution without padding with kernel size 
$3\times3\times X\times19$,
batch-normalization and ReLU activation. It is repeated five times, with $X=64$
the first time and then 19.
Convolution block B comprises of two identical sub-blocks and a skip-connection around the 
second sub-block. Each sub-block consists of a convolution with padding with kernel size
$3\times3\times19\times19$,
batch-normalization and ReLU activation.
The skip-connection adds the output of the first sub-block's ReLU activation to the output
of the second sub-block's batch-normalization. This is repeated twelve times. We found
this architecture to be more effective than a standard residual block \cite{he:16}, since
it achieved $\sim10\%$ better validation accuracy 
at the surrogate task of camera-model identification during training.
The two convolution blocks together learn the spatial distribution of residual values
and can be interpreted as learning their co-occurrences. The final ``bottleneck'' 
convolution has kernel size $3\times3\times19\times1$. 
Its output is a \textit{pre-feature image} of size $56\times56$. All convolutions have stride 1.
Following these are three fully-connected layers: FC1 with 75 neurons, FC2, the 
\textit{feature-layer}, with 100 neurons, and FC3, the final layer that outputs logits, with a 
number of neurons, $C$, corresponding to the number of training camera models. 
FC1 is followed by a dropout layer with keep-probability of 0.8 and ReLU non-linearity.
The network is trained using cross-entropy loss over the training data:
\begin{equation}
\mathcal{L}_{CE} = - \frac{1}{M}\sum_{i=1}^M \mathbf{y}_i\log(\mathbf{\hat{y}}_i),
\label{eq:CE}
\end{equation}
where $\mathbf{y}_i$ is the camera-model label for the $i^{\textrm{th}}$ training 
data point in the mini-batch of length $M$
and $\mathbf{\hat{y}}_i$ is the softmax value computed from the output of FC3.

\noindent
\textbf{Mutual Information based regularization:}
Mutual information (MI) is a popular metric for registering medical images 
since it captures linear and non-linear dependencies between two random variables 
and can effectively compare images of the same body part across
different modalities with different contrasts (\eg MRI, CT, PET) \cite{maes:03}. 
We take advantage of this property of MI to compute the
dependency of the input patch, $\mathbf{P}_i$, 
with the pre-feature image, $\mathbf{p}_i$, which is the output of the final
convolution layer, although they may have different dynamic contrast ranges. 
Given that $\mathbf{p}_i$ is a transformed version of the residuals
computed by the first layer, the dependency primarily reflects the presence of semantic-edges 
in $\mathbf{p}_i$. Therefore, we consider:
\begin{equation}
\mathcal{R}_{MI} = \frac{1}{M}\sum_{i=1}^M \textrm{MI}(\rho(\mathbf{P}_i), \mathbf{p}_i),
\label{eq:MI}
\end{equation}
as a regularization, 
where $\rho(\cdot)$ allows to approximate MI numerically and is described below.

The complete loss function for training our network combines these various components
and also includes $l_2$ regularization of all weights, $\mathbf{W}$, of the network:
\begin{equation}
\mathcal{L} = \mathcal{L}_{CE} + \lambda\mathcal{R}_{RF} + \gamma\mathcal{R}_{MI}
             + \omega||\mathbf{W}||_2,
\label{eq:costfun}
\end{equation}
where $\lambda$, $\gamma$ and $\omega$ balance the amount of RF constraint
penalty and MI \& $l_2$ regularizations to apply along with the main loss.

\noindent
\textbf{Splice localization:} 
We assume that that genuine part of the image comes from a single camera-model and
has the largest number of pixels, while the spliced region(s) is smaller in comparison.
Therefore, we simplify the localization task to a two-class segmentation problem, 
where the distributions of both the classes are approximated by Gaussian distributions 
and the smaller class represents the departure from the feature-statistics of the larger genuine class. 

First, we subdivide the test image into 
$72\times72\times3$ sized patches and compute the feature vector, FC2, for each patch. 
The amount of overlap between neighbouring patches is a hyper-parameter we discuss later. 
Then, we run an expectation-maximization (EM) algorithm to fit a two-component
Gaussian mixture model to the feature-vectors, to segregate the patches into two classes. 
We rerun this fitting one hundred times
with random initializations and select the solution with the highest likelihood.
This probability map is first ``cleaned'' of spurious noise using morphological 
opening (or closing) operation using a fixed disk of size two. 
Then it is upsampled to the original image's dimensions and used for localizing the tampered region(s). 

\begin{table}[t!]
    \centering
    \begin{tabular}{lcl}
    \hline
    Dataset &  \#Img. & Format \\ \hline \hline
    DSO-1 \cite{carvalho:13} & 100 & PNG \\
    NC16 \cite{medifor:17} & 564 & JPEG (mostly) \\
    NC17-dev1 \cite{medifor:17} & 1191 & JPEG (mostly)\\
    \hline
    \end{tabular}
\caption{Details of test datasets we consider.}
\label{table:data}
\end{table}

\subsection{Implementation Details}

\noindent
\textbf{Training:} 
We trained our network using the Dresden Image Database (B) \cite{gloe:10}, which comprises
of $C=27$ camera-models and almost 17,000 JPEG images. We did not segregate the images by 
their compression quality-factors as we considered these to be part of the camera models
signature. For each camera-model we randomly selected $0.2\%$ and $0.1\%$ of the images as 
validation and test sets, while the remaining files were used for training.
The training comprised of a mini-batch size of $M=50$ patches and 100,000 patches per epoch chosen randomly every epoch.
The network was trained for 130 epochs, using Adam optimizer with a constant learning rate
of $1e-4$ for 80 epochs and then decaying exponentially by a factor of 0.9 over the 
remaining epochs. This took approximately two days on an NVIDIA GTX 1080Ti GPU
for our TensorFlow based implementation.
We obtained optimal results of $\sim72\%$ camera-model identification accuracy on the
validation and test sets for weights (Eq.~\ref{eq:costfun}): 
$\lambda=\gamma=1$ and $\omega=5e-4$, which were found empirically.

\noindent
\textbf{MI:}
We computed the MI in Eq.~\ref{eq:MI} numerically by approximating 
$p(\rho(\mathbf{P}_i)),~p(\mathbf{p}_i)$ and $p(\rho(\mathbf{P}_i),\mathbf{p}_i)$
the marginal and joint distributions of $\mathbf{P}_i$ and $\mathbf{p}_i$,  using histograms (50 bins). 
To do this, we defined $\rho(\cdot)$ as a transform that first converts
$\mathbf{P}_i$ $(72\times72\times3)$ to its gray-scalar version then 
resizes it to the dimensions of $\mathbf{p}_i$ $(56\times56)$.
$\rho(\cdot)$ conserves the semantic-edges in $\mathbf{P}_i$ and aligns them
to the edges in $\mathbf{p}_i$.
Histogram based MI computation is a common approximation
that is widely used in medical imaging \cite{maes:03}.
However, it is also computationally inefficient, which explains the 
long training time. 

\section{Results}

We now demonstrate our proposed method for blind splice detection. 
To evaluate its performance quantitatively, we conduct experiments on three datasets, use
three pixel-level scoring metrics, and compare against two top performing splice detection
algorithms. Additionally, we also present the results of a hyper-parameter search to decide on
the optimal overlap of patches during inference (splice localization).

The datasets we select are DSO-1, NC16 and NC17-dev1 (Table~\ref{table:data}). 
These recent datasets contain realistic manipulations that are challenging to detect.
DSO-1 contains splicing manipulations, where human figures, in whole
or in parts, have been inserted into images of other people.
NC16 and NC17-dev1 are more complex and challenging datasets. Images from these may contain
a series of manipulations that may span the entire image or a relatively small region. 
Furthermore, some of these manipulations may be post-processing operations that are meant 
to make forgery detection more difficult. All three datasets provide binary ground-truth
manipulation masks.

To evaluate the performance quantitatively we consider:
F1 score, Matthews Correlation Coefficient (MCC) and area under the receiver operating 
characteristic curve (ROC-AUC).
These metrics have been adopted widely by the digital image forensics community 
\cite{zampoglou:16,cozzolino:18b}.
Since our proposed method generates a probability map, F1 and MCC require a threshold 
to compute a pixel-level binary mask. Again, as per common practice, we report the values
of these scores for the optimal threshold, which is computed with reference to the ground-truth 
manipulation mask \cite{zhou:18,salloum:18,cozzolino:18b}.

\begin{table}[t!]
    \centering
    \begin{tabular}{clll}
    \hline
    Step \qquad &  F1 \qquad \qquad & MCC \qquad \qquad & ROC \qquad \qquad \\
    (pixels) & & & AUC \\ \hline\hline 
    24 & 0.59 & 0.53 & 0.85 \\ 
    36 & 0.65 & 0.61 & 0.89 \\ 
    \textbf{48} & \textbf{0.69} & \textbf{0.65} & \textbf{0.91} \\ 
    60 & 0.68 & 0.64 & \textbf{0.91} \\ 
    72 & 0.67 & 0.64 & 0.90 \\ 
    \hline
    \end{tabular}
\caption{Overlap hyper-parameter search on DSO-1. 
        Best results are achieved for a step of 48 pixels.}
\label{table:GS-DSO1}
\end{table}

\begin{table}[t]
    \centering
    \begin{tabular}{cllll}
    \hline
   Step \qquad  &  F1 \qquad \qquad & MCC \qquad \qquad & ROC \qquad \qquad \\
    (pixels) & & & AUC \\ \hline\hline 
    24 & 0.18 & 0.12 & 0.64 \\
    36 & 0.19 & 0.13 & 0.65 \\ 
    \textbf{48} & \textbf{0.45} & \textbf{0.41} & \textbf{0.81} \\ 
    60 & 0.4 & 0.36 & 0.78 \\ 
    72 & 0.22 & 0.17 & 0.67 \\ 
    \hline
    \end{tabular}
\caption{Overlap hyper-parameter search on 100 randomly selected test images from NC16. 
        Best results are achieved for a step of 48 pixels.}
\label{table:GS-NC16}
\end{table}

\begin{table}[t]
    \centering
    \begin{tabular}{cllll}
    \hline
    Step \qquad  &  F1 \qquad \qquad & MCC \qquad \qquad & ROC \qquad \qquad \\ 
    (pixels) & & & AUC \\ \hline\hline 
    24 & 0.33 & 0.17 & 0.70 \\ 
    36 & 0.34 & 0.19 & 0.71 \\ 
    \textbf{48} & \textbf{0.38} & 0.22 & 0.73 \\ 
    60 & 0.36 & 0.22 & 0.73 \\ 
    72 & 0.36 & \textbf{0.23} & \textbf{0.74} \\ 
    \hline
    \end{tabular}
\caption{Overlap hyper-parameter search on 100 randomly selected test images from NC17-dev1.
        Results achieved for a step of 48 pixels are comparable to the best results.}
\label{table:GS-NC17}
\end{table}

We compare our approach with two state-of-the-art algorithms: SB and EXIF-SC. 
SB \cite{cozzolino:15}, as discussed above, uses the co-occurrences 
of a residual computed from a single
hand-engineered RF and EM algorithm for splice localization. 
It is also a blind approach which has proven 
its merit as a top performer in the 2017 Nimble Challenge. EXIF-SC \cite{huh:18}, 
is a recent publication that has demonstrated promising potential by applying 
a deep neural network to detect splices by predicting meta-data inconsistency.
For each of these methods we report the scores that we computed in our experiments,
using the original codes/models of the 
authors,\footnote{\url{http://www.grip.unina.it/research/83-image-forensics/100-splicebuster.html}, \\
\url{https://minyoungg.github.io/selfconsistency/}}
along with the scores reported by the authors. 

\begin{figure*}[t!]
\begin{center}
   \includegraphics[width=0.8\linewidth]{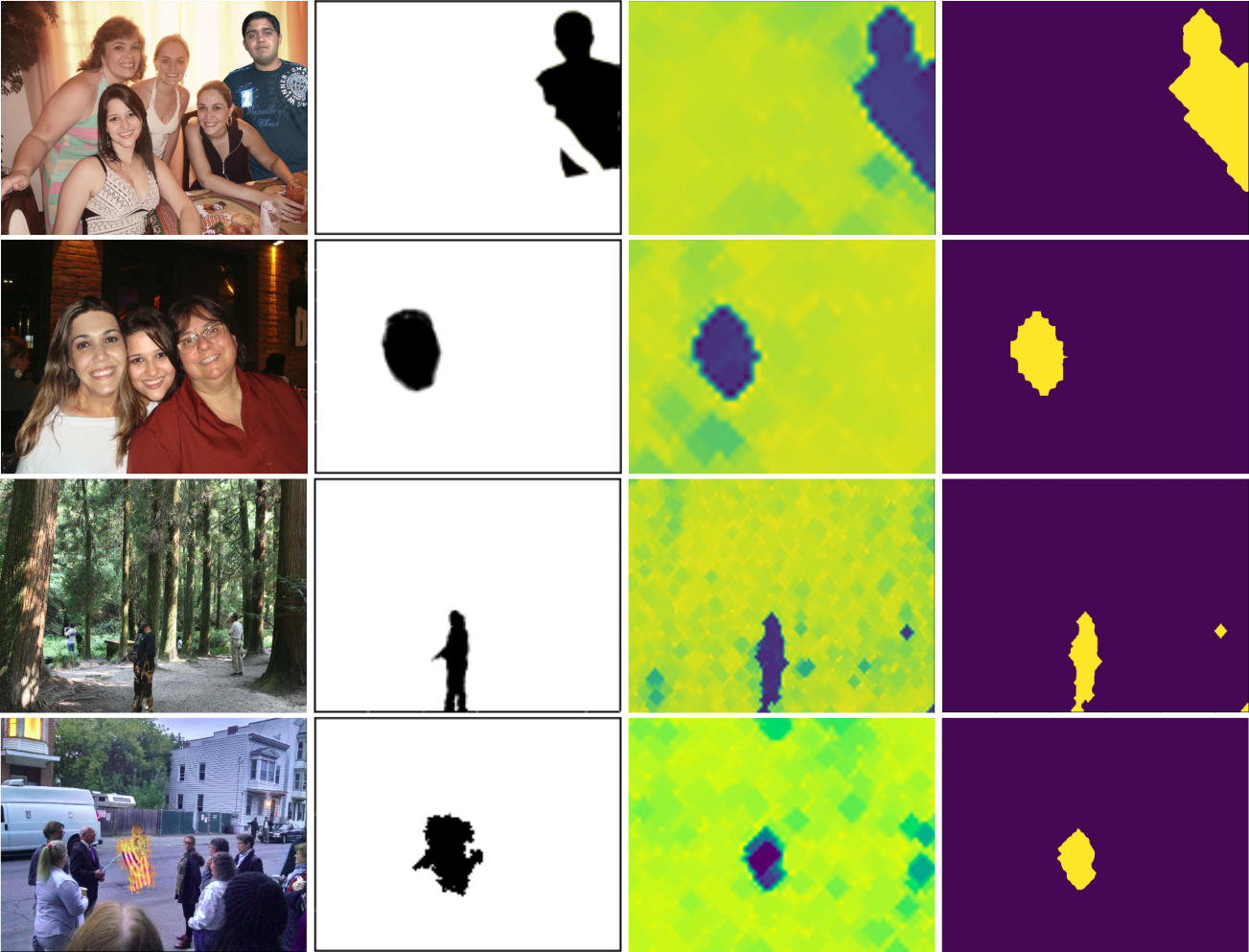}
\end{center}
   \caption{Qualitative results from {SpliceRadar}.
   Col-1: input image, col-2: ground-truth manipulation mask,
   col-3: predicted probability heat map, col-4: predicted binary mask.
   Rows-1,2: DSO-1, row-3: NC16, row-4: NC17-dev1.}
\label{fig:qual}
\end{figure*}

First, we present the results of the hyper-parameter search to decide the optimal overlap of patches
during inference. The overlap is computed in terms of pixels we step along an axis to move from
one patch to the next. 
We compute the performance of our model for steps ranging from 24 to 72 pixels
on the hundred images of DSO-1 and hundred random images of NC16 and NC17-dev1 each. 
The results are presented in Tables~\ref{table:GS-DSO1},\ref{table:GS-NC16},\ref{table:GS-NC17}.
From these we see that a step of 48 pixels produces favourable results consistently. 
Therefore, we consider 48 pixels as the optimal step size in all our experiments.

Next, we present the results of forgery detection. 
Table~\ref{table:F1} presents the F1 scores
achieved by all three algorithms over the three test datasets.
SpliceRadar is able to improve over the performances of SB and EXIF-SC
on DSO-1 and NC16, while its performance is on par with them on NC17-dev1.
Table~\ref{table:MCC} presents the MCC results in a similar format. 
Again, SpliceRadar outperforms SB and EXIF-SC 
on DSO-1 and NC16 and ties with
SB as a top performer on NC17-dev1.
The ROC-AUC results are presented in Table~\ref{table:AUC}.
In this case, SpliceRadar has the best scores on all three datasets,
indicating a better global performance across all thresholds.
Overall, from these three tables, we observe that our proposed 
method's performance
is not only comparable to the state-of-the-art, 
but up to $4\%$ points better.

\begin{table}[t!]
    \centering
    \begin{tabular}{llll}
    \hline
     &  DSO-1 & NC16 & NC17-dev1 \\ \hline \hline
    EXIF-SC & 0.57 \src{0.52} & 0.38 & 0.41 \\
    SB &  0.66 \src{0.66} & 0.37 \src{0.36} & \textbf{0.43} \\
    \textbf{SR} & \textbf{0.69} & \textbf{0.40} & 0.42 \\
    \hline
    \end{tabular}
\caption{Results: F1 score comparison on the test datasets. Black: scores we computed, 
        blue: scores reported by the authors. (For SB, we cite results from \cite{cozzolino:18b}).}
\label{table:F1}
\end{table}

\begin{figure*}[t!]
\begin{center}
   \includegraphics[width=0.8\linewidth]{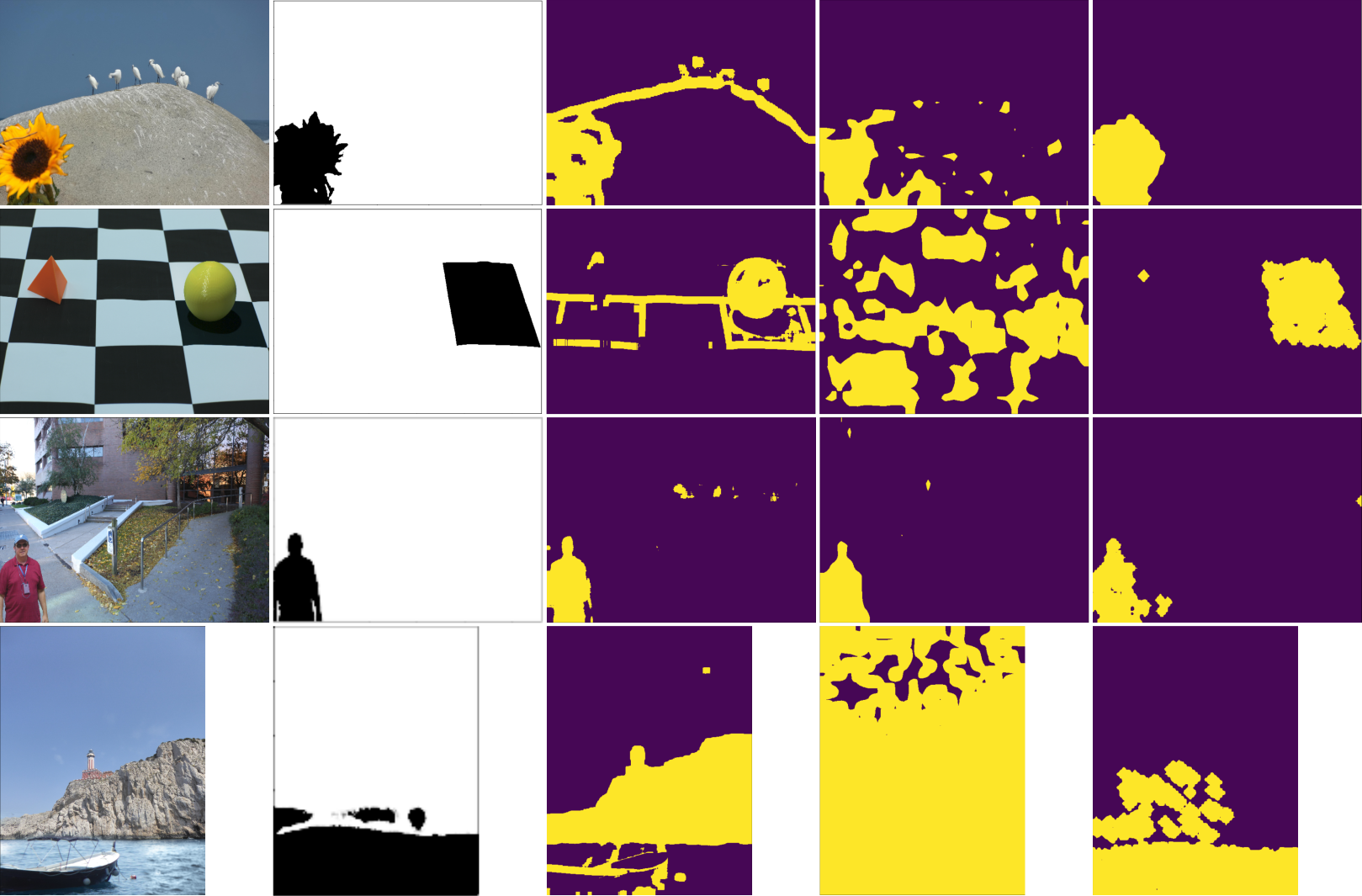}
\end{center}
   \caption{Qualitative comparison of SpliceRadar, SB and EXIF-SC.
   Col-1: input image, col-2: ground-truth manipulation mask,
   col-3: mask from SB, col-4: mask from EXIF-SC,
   col-5: mask from SpliceRadar.
   Rows-1,2: NC16, rows-3,4: NC17-dev1.}
\label{fig:comp}
\end{figure*}

We present qualitative results in Figs.~\ref{fig:qual},\ref{fig:comp}, where
we select examples from all three datasets DSO-1, NC16 and NC17-dev1.
Fig.~\ref{fig:qual} shows the input colour image in the first column, 
the ground-truth manipulation mask in the second column, 
the probability heat map predicted by SpliceRadar in the third column 
and the predicted binarized manipulation mask in the final column.
In Fig.~\ref{fig:comp}, we qualitatively compare the 
predicted binarized masks
of all three algorithms compared in Tables~\ref{table:F1},\ref{table:MCC},\ref{table:AUC} 
alongside the input image and the ground-truth manipulation mask. 
These figures provide a visual insight into our method's performance.

\begin{figure*}
\begin{center}
   \includegraphics[width=0.8\linewidth]{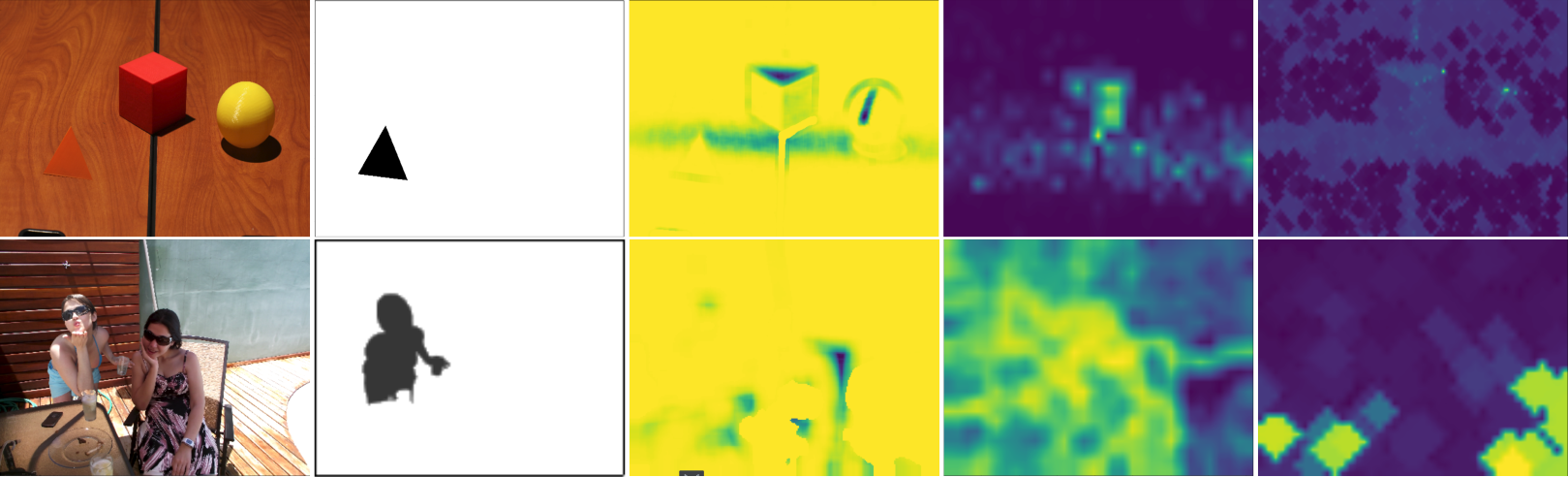}
\end{center}
   \caption{Hard examples where all three algorithms, 
   SpliceRadar, SB and EXIF-SC, fail to detect the spliced regions.
   Col-1: input image, col-2: ground-truth manipulation mask,
   col-3: heat map from SB, col-4: heat map from EXIF-SC,
   col-5: heat map from SpliceRadar.   }
\label{fig:fail}
\end{figure*}

Finally, in Fig.~\ref{fig:fail} we present some hard examples,
where all three algorithms fail to detect the spliced regions.
These examples require further investigation and indicate future research directions.

\begin{table}
    \centering
    \begin{tabular}{llll}
    \hline
     &  DSO-1 & NC16 & NC17-dev1 \\ \hline \hline
    EXIF-SC & 0.52 \src{0.42} & 0.36 & 0.18 \\
    SB &  0.61 \src{0.61} & 0.34 \src{0.34} & \textbf{0.2} \\
    \textbf{SR} & \textbf{0.65} & \textbf{0.38} & \textbf{0.2} \\
    \hline
    \end{tabular}
\caption{Results: MCC score comparison on the test datasets. Black: scores we computed, 
        blue: scores reported by the authors. (For SB, we cite results from \cite{cozzolino:18b}).}
\label{table:MCC}
\end{table}

\begin{table}
    \centering
    \begin{tabular}{llll}
    \hline
     &  DSO-1 & NC16 & NC17-dev1 \\ \hline \hline
    EXIF-SC & 0.85 \qquad\qquad & 0.80 \qquad\qquad & 0.71 \\
    SB &  0.86  & 0.77  & 0.69 \\
    \textbf{SR} & \textbf{0.91} & \textbf{0.81} & \textbf{0.73} \\
    \hline
    \end{tabular}
\caption{Results: ROC-AUC score comparison on the test datasets. }
\label{table:AUC}
\end{table}


\section{Conclusion and Future Directions}

We proposed a novel method for blind forgery localization using 
a deep convolutional neural network that learns low-level features 
capable of segregating camera-models.
These low-level features, independent
of the semantic contents of the training images, were learned in two 
stages: first, using our new constrained convolution approach to learn
relevant residuals and second, using our novel probabilistic MI-based
regularization to suppress semantic-edges. Preliminary results on three
test datasets demonstrated the potential of our approach, indicating up 
to $4\%$ points improvement over the state-of-the-art.

In this first study, we compared our approach with two   top performing
state-of-the-art methods on three datasets.
We plan more extensive tests in the future with more 
recent datasets like those from 
Media Forensics Challenge 2018 and more algorithms. 
We plan to
also systematically investigate the effects of JPEG compression.

One shortcoming of our approach is the histogram based implementation of
mutual information, which is computationally cumbersome. 
This compelled us to curtail our model in a number of ways:
to use a relatively small mini-batch size, 
to train for a limited number of 
epochs and to consider a relatively small network. We plan to improve this
bottleneck in the future to enable us to train larger models on bigger
datasets more efficiently. 
We also identified hard examples where all the algorithms we tested failed
to identify the correct spliced regions. These require further investigation.
Finally, we foresee including more prior 
knowledge to improve results, for example fine-tuning our model
on the training data provided with each dataset.


{\small
\bibliographystyle{ieee}
\bibliography{forensics}
}

    \end{document}